\documentclass[conference]{IEEEtran}
\IEEEoverridecommandlockouts
\usepackage{times}
\usepackage{soul}
\usepackage{url}
\usepackage[hidelinks]{hyperref}
\usepackage[utf8]{inputenc}
\usepackage[small]{caption}
\usepackage{graphicx}
\usepackage{amsmath}
\usepackage{amsthm}
\usepackage{booktabs}
\usepackage{algorithm}
\usepackage{algorithmic}
\usepackage[switch]{lineno}

\usepackage{float}
\usepackage{listings}
\usepackage{amssymb}
\usepackage{amsfonts}
\usepackage{bm}
\usepackage{multicol}
\usepackage{subfigure}
\usepackage{url}
\usepackage{booktabs} 
\usepackage{color}
\usepackage{arydshln} 

\usepackage{bm}
\usepackage{stfloats}
\def\BibTeX{{\rm B\kern-.05em{\sc i\kern-.025em b}\kern-.08em
    T\kern-.1667em\lower.7ex\hbox{E}\kern-.125emX}}
\begin{document}

\title{Concept Learning for Cooperative Multi-Agent Reinforcement Learning\\

\thanks{*Corresponding Author.\\ This work was supported in part by the China Postdoctoral Science Foundation under Grant Number 2025T180877, the National Key Research and Development Program of China under Grant 2023YFD2001000 and the Nanjing University Integrated Research Platform of the Ministry of Education-Top Talents Program.}
}

\author{\IEEEauthorblockN{Zhonghan Ge}
\IEEEauthorblockA{\textit{School of Mangement Engineering} \\
\textit{Nanjing University}\\
Nanjing, China \\
1476360984@qq.com}
\and
\IEEEauthorblockN{Yuanyang Zhu\textsuperscript{*}}
\IEEEauthorblockA{\textit{School of Information Mangement} \\
\textit{Nanjing University}\\
Nanjing, China \\
yuanyangzhu@nju.edu.cn}
\and
\IEEEauthorblockN{Chunlin Chen}
\IEEEauthorblockA{\textit{School of Mangement Engineering} \\
\textit{Nanjing University}\\
Nanjing, China \\
clchen@nju.edu.cn}}

\maketitle

\begin{abstract}
Despite substantial progress in applying neural networks (NN) to multi-agent reinforcement learning (MARL) areas, they still largely suffer from a lack of transparency and interpretability.
However, its implicit cooperative mechanism is not yet fully understood due to black-box networks.
In this work, we study an interpretable value decomposition framework via concept bottleneck models, which promote trustworthiness by conditioning credit assignment on an intermediate level of human-like cooperation concepts.
To address this problem, we propose a novel value-based method, named Concepts learning for Multi-agent Q-learning (CMQ), that goes beyond the current performance-vs-interpretability trade-off by learning interpretable cooperation concepts.
CMQ represents each cooperation concept as a supervised vector, as opposed to existing models where the information flowing through their end-to-end mechanism is concept-agnostic.
Intuitively, using individual action value conditioning on global state embeddings to represent each concept allows for extra cooperation representation capacity.
Empirical evaluations on the StarCraft II micromanagement challenge and level-based foraging (LBF) show that CMQ achieves superior performance compared with the state-of-the-art counterparts.
The results also demonstrate that CMQ provides more cooperation concept representation capturing meaningful cooperation modes, and supports test-time concept interventions for detecting potential biases of cooperation mode and identifying spurious artifacts that impact cooperation.
\end{abstract}


\section{Introduction}
Cooperative multi-agent reinforcement learning (MARL) has achieved notable success in complex domains such as autonomous driving~\cite{car}, sensor networks~\cite{zhang2011coordinated}, and robotics control~\cite{huttenrauch2017guided}.
A core driver behind this progress is the development of value decomposition methods~\cite{vdn,qmix,qplex,qtran,qpd}, which effectively factorize the joint action-value function to enable decentralized execution with centralized training.
However, these methods often operate as black boxes, providing limited visibility into individual agent behaviors or their contributions to team performance, hindering transparency, trust, and practical deployment in safety-critical domains.

Efforts to enhance interpretability in RL typically fall into post-hoc analysis or global approximation~\cite{mixrts,na2q}.
Instance-based methods (e.g., Shapley values~\cite{wang2020shapley}, clustering~\cite{zahavy2016graying}) offer local explanations but suffer from instability and computational cost~\cite{ghorbani2019interpretation,slack2021reliable}.
Global imitation-based techniques~\cite{bastani2018verifiable,silva2020optimization,9970401} distill agent behaviors but often fail to retain fidelity in complex tasks.
Recently, concept-based learning~\cite{koh2020concept,stammer2022interactive,havasi2022addressing} has emerged as a promising paradigm, offering high-level, human-understandable representations of a model's reasoning process.
Concept bottleneck models, in particular, separate learning into two stages: predicting intermediate concepts, then making final decisions based on these concepts, enabling both interpretability and test-time human intervention~\cite{bai2022concept}.

Inspired by this, we explore whether concept learning can enhance interpretability and coordination in MARL. We propose CMQ (Concept learning for Multi-agent Q-learning), a novel value decomposition method that integrates rich cooperation concepts as intermediate representations. 
Each agent's local action-value is combined with a global concept embedding to compute the joint action-value and agent-wise credit.
The concept bottleneck not only improves interpretability but also retains expressiveness through concept-conditioned temporal Q-values, mitigating the linearity limitations in existing decomposition methods~\cite{qplex}.
Furthermore, CMQ supports test-time concept interventions, allowing practitioners to diagnose and adjust specific cooperation failures.

Our contributions are summarized as follows: 1) We propose a novel value decomposition method, called Concept learning for Multi-agent Q-learning (CMQ), which moves a step toward modeling explicit collaboration among agents;
2) We provide the test-time concept intervention to diagnose which cooperation concepts are incorrect or do not align with human experts;
3) Through extensive experiments on challenging MARL benchmarks, CMQ not only consistently achieves superior performance compared to different state-of-the-art baselines but also allows for an easy-to-understand interpretation of credit assignment.

\section{Preliminaries and Related Work}
\textbf{Dec-POMDP.}
Cooperative multi-agent tasks are generally formulated as decentralized partially observable Markov decision processes (Dec-POMDPs), defined by the tuple $G = \langle \mathcal{N}, \mathcal{S}, \mathcal{A}, \mathcal{P}, \Omega, O, r, \gamma \rangle$, where $\mathcal{N}$ is the agent set, $s \in \mathcal{S}$ is the global state, $\gamma$ is the discount factor, and $a_i \in \mathcal{A}$ is the action of agent $i$.
At each timestep, agents take actions $\boldsymbol{a}$ and receive a shared reward $r(s, \boldsymbol{a})$ with state transitions governed by $\mathcal{P}$.
Under partial observability, each agent receives local observation $o_i$ from $O(o_i | s, a_i)$ and forms a history $\tau_i$.
The goal is to learn a joint policy $\boldsymbol{\pi} = \langle \pi_1, \ldots, \pi_n \rangle$ that maximizes the expected cumulative return $R_t = \sum_{t=0}^\infty \gamma^t r_t$.

\textbf{Centralized Training with Decentralized Execution.} CTDE is a prevalent paradigm in the MARL, where each agent learns its policy solely from its own action–observation history, while a centralized critic exploits the full joint state–action information to compute gradient updates. 
A natural extension of this framework is value decomposition, in which agents learn individual utility functions whose combination reconstructs the global action‐value function.
This decomposition not only facilitates efficient credit assignment, but—if properly constrained—also preserves optimality guarantees.
In particular, any value decomposition scheme must satisfy the \emph{Individual–Global–Max} (IGM) principle~\cite{qtran}, which demands that the joint action maximizing the total value coincides with the tuple of per‐agent maximizers:
\begin{equation}
\label{dsdsadad}
\operatorname{argmax}_{\boldsymbol{a}} Q_{tot}(\boldsymbol{\tau}, \boldsymbol{a})
=\left(\begin{array}{c}
\operatorname{argmax}_{a_{1}} Q_{1}\left(\tau_{1}, a_{1}\right) \\
\vdots \\
\operatorname{argmax}_{a_{n}} Q_{n}\left(\tau_{n}, a_{n}\right)
\end{array}\right),
\end{equation}
where $\boldsymbol{a}$ is the joint action-observation history.

\textbf{Concept Bottleneck Models.}
Concept Bottleneck Models (CBMs) structure the learning process into two stages: concept learning and decision-making.
Given an input $x$, a CBM first maps $x$ to an intermediate representation $c$, which encodes a set of predefined concepts.
These concepts are chosen for their interpretability and relevance to the task.
The model then uses $c$ to make a final prediction $y$.
Formally, a CBM can be represented as a composite function $y = g(f(x))$, where $f: \mathcal{X} \rightarrow \mathcal{C}$ is the concept learning function and $g: \mathcal{C} \rightarrow \mathcal{Y}$ is the decision-making function.
The overall objective in training a CBM is to optimize both $f$ and $g$ to improve the interpretability of the model while maintaining or even enhancing its predictive performance.

\textbf{Related work.}
Effective credit assignment is at the core of value‐decomposition methods for cooperative MARL, as it allocates the global action‐value among agents to guide decentralized policy updates~\cite{vdn,qmix}.
VDN~\cite{vdn} achieves this via a simple linear sum of individual Q‐values, while QMIX~\cite{qmix} enforces a monotonic mixing network to guarantee consistency with the IGM principle.
Building on this foundation, QTRAN~\cite{qtran} relaxes the monotonicity constraint through specialized transformation objectives, QPLEX~\cite{qplex} introduces a dueling‐style non‐monotonic mixer, and both WQMIX~\cite{rashid2020wqmix} and Qatten~\cite{qatten} enrich the mixer with distributional and attention‐based mechanisms for finer credit estimation.
Despite these advances, the mixing networks in such approaches remain largely opaque and ignore the underlying semantic factors that drive effective coordination.  
Alternatives, QPD~\cite{qpd} computes credit weights directly from each agent's action–observation history, but their black-box mixers still lack transparency.
In contrast, we propose to embed concept learning into the decomposition process, thereby structuring the joint value function around human‐interpretable semantic units.
This not only preserves strong cooperative performance but also yields transparent credit assignments that align with interpretable concepts, closing the gap between high-performance and interpretability.

\begin{figure*}[tb]
   \centering
    \includegraphics[width=2\columnwidth]{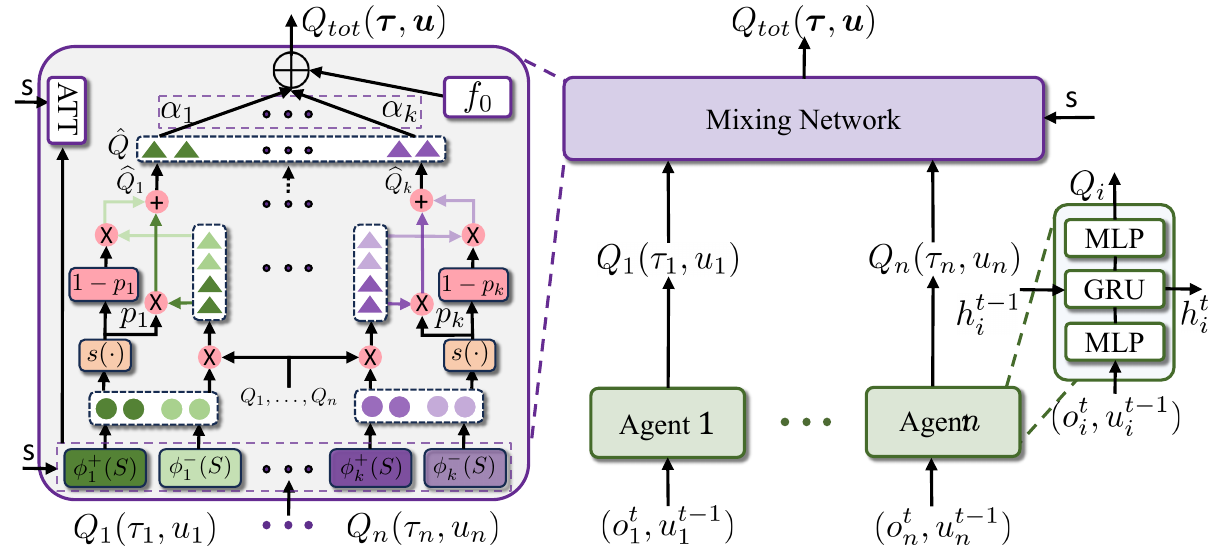}
    \caption{The framework of our method CMQ. 
    First, each agent models a value function $Q_i(\tau_i, a_i)$ conditioned on its individual action-observation history $\tau_i$.
To model the activation of these concepts, we parameterize each one with a pair of global-state-dependent embeddings: $\hat{\mathbf{s}}_i^{+}$ when the concept is active, and $\hat{\mathbf{s}}_i^{-}$ when inactive.
These embeddings are aligned with agent-level Q-values via dot-product interactions, serving as semantic interpreters of cooperation patterns.
A learnable scoring function $s(\cdot)$ estimates the activation probability $\hat{p}_i$ for each concept based on its dual embeddings.
These probabilities are used to interpolate between the active and inactive representations, yielding a fused embedding for each concept: $\widehat{Q}_i = \hat{p}_i \cdot \widetilde{Q}_i^{+} + (1 - \hat{p}_i) \cdot \widetilde{Q}_i^{-}$.
This formulation enables concept-aware credit assignment while preserving semantic interpretability and enforcing structural disentanglement between latent cooperation modes.
    }
    \skip -0.2in
    \label{framework}
\end{figure*}

\section{Concept Learning for Cooperation MARL}
To overcome the limitations of existing value decomposition methods, we introduce Concept-based Multi-agent Q-learning (CMQ)—a novel framework that integrates concept bottleneck learning into value decomposition, aiming to make multi-agent cooperation both effective and interpretable.
CMQ factorizes the global action-value function as a weighted sum over concept-conditioned temporal Q-values:

\begin{equation}
\label{eq1}
Q_{tot}^\phi(\boldsymbol{\tau}, \mathbf{a}) = \sum_k \alpha_k \widehat{Q}_k(\boldsymbol{\tau}, a_i) + f(\boldsymbol{s}),
\end{equation}
where $\widehat{Q}_k(\boldsymbol{\tau}, a_i)$ denotes the concept-specific temporal Q-value generated via agent activation probabilities $p_i$ and individual value components, while $\alpha_k$ is a state-dependent weight reflecting global semantics.
This decomposition not only enhances transparency through interpretable cooperation concepts but also retains policy learning effectiveness.
Although linear decomposition may be limited in expressiveness~\cite{qplex}, we show that CMQ maintains competitive performance without sacrificing interpretability.

As illustrated in Figure~\ref{framework}, CMQ consists of local agent value functions $Q_i(\tau_i, u_i)$, which are aggregated through a concept-driven mixing network.
Local Q-values $[Q_i]_{i=1}^n$ are mapped into cooperation concept bottlenecks $\widehat{Q}_i$, and global state embeddings are used to compute corresponding credit assignments.
The final joint Q-value is computed from these concept-conditioned values and their associated credits.
Importantly, CMQ introduces explicit cooperation semantics, enabling test-time concept interventions to simulate how specific modes of coordination impact performance.

In the following, we elaborate on the structure of the mixing network, how it integrates dual semantic embeddings per concept, and how CMQ enables meaningful concept-level interventions, making it responsive and reliable in real-world cooperative scenarios.

\subsection{Architecture of CMQ}
CMQ comprises a concept predictor and a joint value function predictor.
The concept predictor generates a mixture of two global state semantics for each cooperation concept to explicitly predict the probability of the concept's activity.
The joint value function predictor estimates the joint value from the predicted concept embeddings together with the global state $s$ and the cooperation concept bottleneck $\widehat{Q}_{i}$.
This architecture endows our model with the capability to assemble arguments both supporting and contesting the activation of a concept, thereby facilitating straightforward interventions in cooperation concepts.
At the point of intervention, one can deftly alternate between the global state semantics, thereby tailoring concept activation dynamics.

In practice, we use the concept predictor to produce cooperation concept $c_i(\mathbf{s})$ with two global state semantics $\hat{c}_i^{+}(\mathbf{s}), \hat{c}_i^{-}(\mathbf{s}) \in \mathbb{S}^m$.
We define the specific semantics: $\hat{c}_i^{+}(\mathbf{s})$ and $\hat{c}_i^{-}(\mathbf{s})$, which represent the existence and non-existence of concept $c_i(\mathbf{s})$, respectively.
To achieve this, the global state  $s$ is fed into two concept-specific fully connected layers, yielding the following embeddings $\hat{c}_i^{+}(\mathbf{s})=\phi_i^{+}(\mathbf{s})=Relu(W_i^{+} \mathbf{s}+\mathbf{b}_i^{+})$ and $\hat{c}_i^{-}(\mathbf{s})=\phi_i^{-}(\mathbf{s})=$ $Relu((W_i^{-} \mathbf{s}+\mathbf{b}_i^{-})$, where $W_i^{+}, W_i^{-}$ and $\mathbf{b}_i^{+}, \mathbf{b}_i^{-}$ are learnable parameters.
These dual embeddings provide a semantic grounding that distinguishes whether the corresponding concept is active or inactive under a given global state.
To align the embeddings with the underlying ground-truth semantics, we employ a learnable and differentiable scoring function $s$: $\mathbb{S}^{2m}\rightarrow[0,1]$, which is adeptly crafted to estimate the probability $\hat{p}_i \triangleq s(\left[\hat{c}_i^{+}(\mathbf{s}), \hat{c}_i^{-}(\mathbf{s})\right]^T)=\sigma(W_s\left[\hat{c}_i^{+}(\mathbf{s}), \hat{c}_i^{-}(\mathbf{s})\right]^T+\mathbf{b}_\mathbf{s})$ of concept $c_i(\mathbf{s})$ is in existence, based on the conjoint embedding space, where $\sigma$ is a sigmoid function.
To promote parameter efficiency and generalization, the scoring function parameters $W_s$ and $\mathbf{b}_s$ are shared across all concepts $c_i(\mathbf{s})$, enabling compact modeling of semantic decision boundaries across multiple concept dimensions.

To relieve the limited expressivity of linear decomposition for the joint value function, we factor the centralized critic as a weighted summation of individual critics across agents.
We first build the temporal Q-values $\widetilde{Q}_{i}$ that comprise the positive $\widetilde{Q}_{i}^{+}= W_i^{+}(\mathbf{s}) [Q_1\left(\boldsymbol{\tau}, \mu\right),...,Q_n\left(\boldsymbol{\tau}, \mu\right)]$ and negitive $\widetilde{Q}_{i}^{-}= W_i^{-}(\mathbf{s}) [Q_1\left(\boldsymbol{\tau}, \mu\right),...,Q_n\left(\boldsymbol{\tau}, \mu\right)]$.
Given the concept activation probability $\hat{p}_i$, we compute the final concept-level Q-value via a convex combination of the positive and negative projections:
\begin{equation}
\widehat{Q}_{i} \triangleq\left(\hat{p}_i \widetilde{Q}_{i}^{+}+\left(1-\hat{p}_i\right) \widetilde{Q}_{i}^{-}\right).
\end{equation}

Intuitively, this serves a two-fold purpose: (i) it enforces selective dependence on $\hat{\mathbf{s}}_i^{+}$ when the concept $i$ is active (i.e., $\hat{p}_i \approx 1$), and on $\widetilde{Q}_{i}^{-}$ otherwise, thereby inducing two semantically meaningful latent spaces, and (ii) it enables a clear intervention strategy where one switches the embedding states when mispredicted concepts can be corrected by swapping their embedding states.
All $k$ mixed concept Q-values $\{\widehat{Q}_k\}_{k=1}^{k}$ are then concatenated to form a cooperation bottleneck.
To realize the full joint value function $Q_{tot}$ in Eq.~\ref{eq1}, we compose it as a weighted sum over the concept-conditioned Q-values:
\begin{equation}
Q_{tot}^\phi(\boldsymbol{\tau}, \mathbf{a})=\sum_k \alpha_k \widehat{Q}_k\left(\boldsymbol{\tau}, a_i\right)+f(\boldsymbol{s}),
\end{equation}
where $f_0(\mathbf{s})$ is a bias term, $\widehat{Q}_k\left(\boldsymbol{\tau}, a_i\right)$ is the cooperation bottleneck and $\alpha{i}$ is credits.
The credit $\alpha_k$  is computed using a dot-product attention mechanism between the concept embedding $\hat{c}_i(\mathbf{s}$ and the global state $s$:
\begin{equation}
\alpha_k=\frac{\exp \left(\left(\boldsymbol{w}_i \hat{c}_i(\mathbf{s})\right)^{\top} \operatorname{ReLU}\left(\boldsymbol{w}_\mathbf{s} \boldsymbol{s}\right)\right)}{\sum_{k=1}^m \exp \left(\left(\boldsymbol{w}_i \hat{c}_i(\mathbf{s})\right)^{\top} \operatorname{ReLU}\left(\boldsymbol{w}_s \boldsymbol{s}\right)\right)},
\end{equation}
where $\boldsymbol{w}_i$ and $\boldsymbol{s}$ are the learnable parameters, and ReLU serves as the activation function to preserve non-negativity.
To ensure the monotonicity constraint of Eq.~(\ref{dsdsadad}), the credits $\alpha_k$ are enforced to be non-negative, e.g., via absolute-value operations if necessary.
Finally, similar to vanilla CBMs, our framework CMQ provides interpretable reasoning via its concept probability vector $\hat{\mathbf{p}}(\mathbf{x}) \triangleq\left[\hat{p}_1, \cdots, \hat{p}_k\right]$, which explicitly indicates the predicted activation of each cooperation concept involved in shaping $Q_{tot}$.

Under the CTDE paradigm, CMQ is optimized by minimizing the standard squared temporal-difference (TD) error.
Specifically, during training, a batch of transitions is sampled from a replay buffer $b$, and the loss is defined as:
\begin{equation}
\mathcal{L}(\theta)=\mathbb{E}_{(\boldsymbol{\tau}, \boldsymbol{a}, r, \boldsymbol{\tau}^{\prime})\in b}\left[\left(y-Q_{tot}(\boldsymbol{\tau}, \boldsymbol{a}; \theta)\right)^{2}\right],
\end{equation}
where $\theta$ denotes the current network parameters, and $\boldsymbol{\tau}$ represent the joint action-observation trajectory.
The TD target value is computed as $y=r+\gamma \max _{\boldsymbol{a}^{\prime}} Q_{tot}\left(\boldsymbol{\tau}^{\prime}, \boldsymbol{a}^{\prime}; \theta^{-}\right)$, where $\theta^{-}$ denotes the parameters of a target network that are periodically updated from $\theta$ and held fixed for a number of training iterations to stabilize learning.
It allows CMQ to learn semantically grounded joint value estimates by integrating concept-based representations with standard RL optimization.

\subsection{Intervening with Cooperation Concept}
In our CMQ framework, concept intervention is straightforward: it entails substituting the predicted cooperation concept with the one that aligns semantically with the ground truth.
For example, at time step $t$ within the context of a given global state $s$, individual action values and concept $\hat{c}_i^{-}(\mathbf{s})$, CMQ predicted $\hat{p}_i=0.1$ while a human expert knows that concept $\hat{c}_i^{-}(\mathbf{s})$ is activate ($\hat{c}_i^{-}(\mathbf{s})=1$), an intervention can be enacted to adjust $\hat{p}_i=1$.
This revision recalibrates the cooperation concept representation within the CMQ bottleneck from a weighted mixture $(0.1 \hat{\mathbf{c}}_i^{+}(\mathbf{s})+0.9 \hat{\mathbf{c}}_i^{-}(\mathbf{s}))$ to the pure concept $\hat{c}_i^{+}$.
Such refinement directly informs the joint value function predictor with the accurate concept context.

Furthermore, a regularization technique is employed during training to acclimate the CMQ model to concept interventions, enhancing its responsiveness during deployment.
This involves executing random concept interventions with a probability $\widetilde{p}$, which can be set to $\hat{p}_i:=c_i(\mathbf{s})$ for the concept $\hat{c}_i^{-}(\mathbf{s})$ with probability $\widetilde{p}$.
The computation of the concept $\hat{c}_i(\mathbf{s})$ during training process is as follows:
\begin{equation}
\hat{\mathbf{c}}_i(s)=
\begin{cases}
c_i(s) \hat{\mathbf{c}}_i^{+}(s)+(1-c_i(s)) \hat{\mathbf{c}}_i^{-}(s), & \text{with } \widetilde{p} \\
\hat{p}_i \hat{\mathbf{c}}_i^{+}(s)+\left(1-\hat{p}_i\right) \hat{\mathbf{c}}_i^{-}(s), & \text{with } 1-\widetilde{p}
\end{cases}
\end{equation}
while the predicted probability mixture is systematically employed at inference time.
During the backpropagation phase, the updating process is selective: only the relevant concept embedding (i.e., $\hat{\mathbf{c}}_i^{+}(\mathbf{s})$ if $\hat{\mathbf{c}}_i(\mathbf{s})=1$) is updated based on task-specific feedback, whereas concept prediction feedback impacts both embeddings $\hat{\mathbf{c}}_i^{+}(\mathbf{s})$ and $\hat{\mathbf{c}}_i^{-}(\mathbf{s})$
Conceptually, this approach encapsulates the learning across a vast array of CMQ configurations, where individual concept representations are either honed by isolated concept feedback or by an amalgamation of concept and task-derived feedback.

\section{Experiments}
In this section, we evaluate the performance of our proposed approach, CMQ, against several strong baselines.
Specifically, we compare with three representative value decomposition methods, VDN~\cite{vdn}, QMIX~\cite{qmix}, and QTRAN~\cite{qtran}, and four state-of-the-art MARL algorithms, including WQMIX~\cite{rashid2020wqmix}(both CW-QMIX and OW-QMIX), QPLEX~\cite{qplex}, CDS~\cite{li2021cds}, and SHAQ~\cite{wang2021shaq}.
We conduct experiments on two widely-used cooperative MARL benchmarks: the Level-Based Foraging (LBF) environment~\cite{christianos2020shared}, and the StarCraft II micromanagement benchmark (SMAC)~\cite{smac}, using SC2 version 2.4.10. 
All experiments are repeated across five random seeds to account for stochasticity, and we report the mean performance along with the 75\% confidence intervals. 
Our results demonstrate that CMQ consistently outperforms all baselines in both final performance and learning efficiency.
We also conduct ablation studies to analyze the effect of varying the number of semantic concepts, providing insight into how concept granularity influences coordination.
Finally, we evaluate the interpretability of CMQ by visualizing semantic contributions and demonstrating how concept interventions affect decision-making behavior.

To ensure fair comparisons, all methods are trained under consistent hyperparameter settings: a batch size of $32$, discount factor $\gamma = 0.99$, and an exploration rate linearly decaying from $1.0$ to $0.05$.
The replay buffer size is set to $5000$ transitions, and the target networks are updated every $200$ episodes.
Optimization is performed using RMSprop with a learning rate of $0.0005$, and gradients are clipped to a maximum norm of $10$.
Our CMQ model is implemented within the PyMARL framework. 
We use $16$ semantic concepts to construct the concept space, and the bias term is modeled via a single-layer MLP with $32$ hidden units. 
All attention modules use a hidden dimension of $64$.
Training is conducted on a single machine equipped with an NVIDIA RTX 4090 GPU and an Intel i9-13900K CPU.
Depending on the scenario complexity, training requires between 2M and 5M timesteps, which corresponds to approximately 0.5 to 5 hours of wall-clock time.

\begin{figure}[h]
	\begin{center}
		\centerline{\includegraphics[width=1\columnwidth]{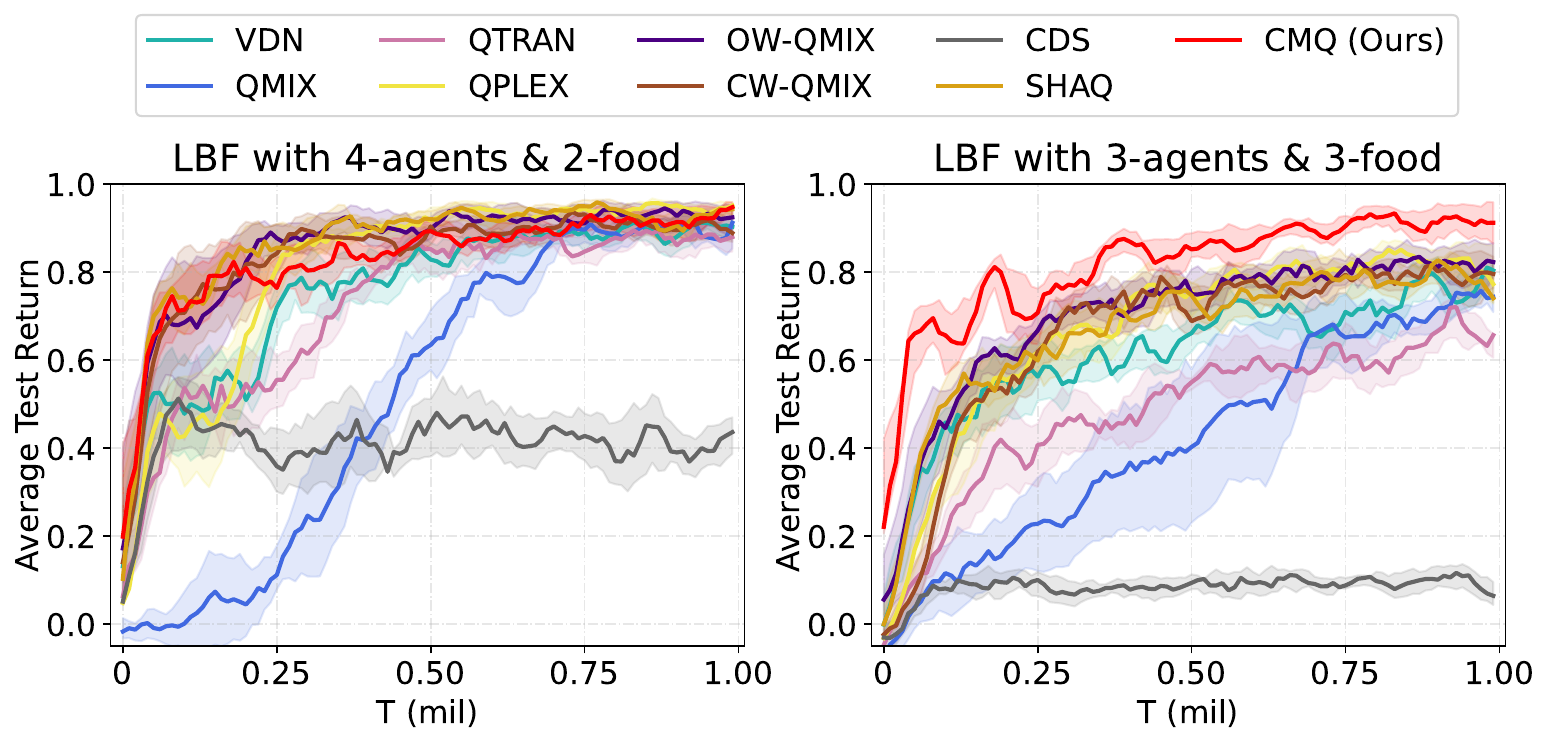}}
		\caption{Average test return on two constructed tasks of LBF.}
		\label{overview_results_lbf}
	\end{center}
	\vskip -0.25in
\end{figure}

\begin{figure*}[tb]
   \centering
    \includegraphics[width=2\columnwidth]{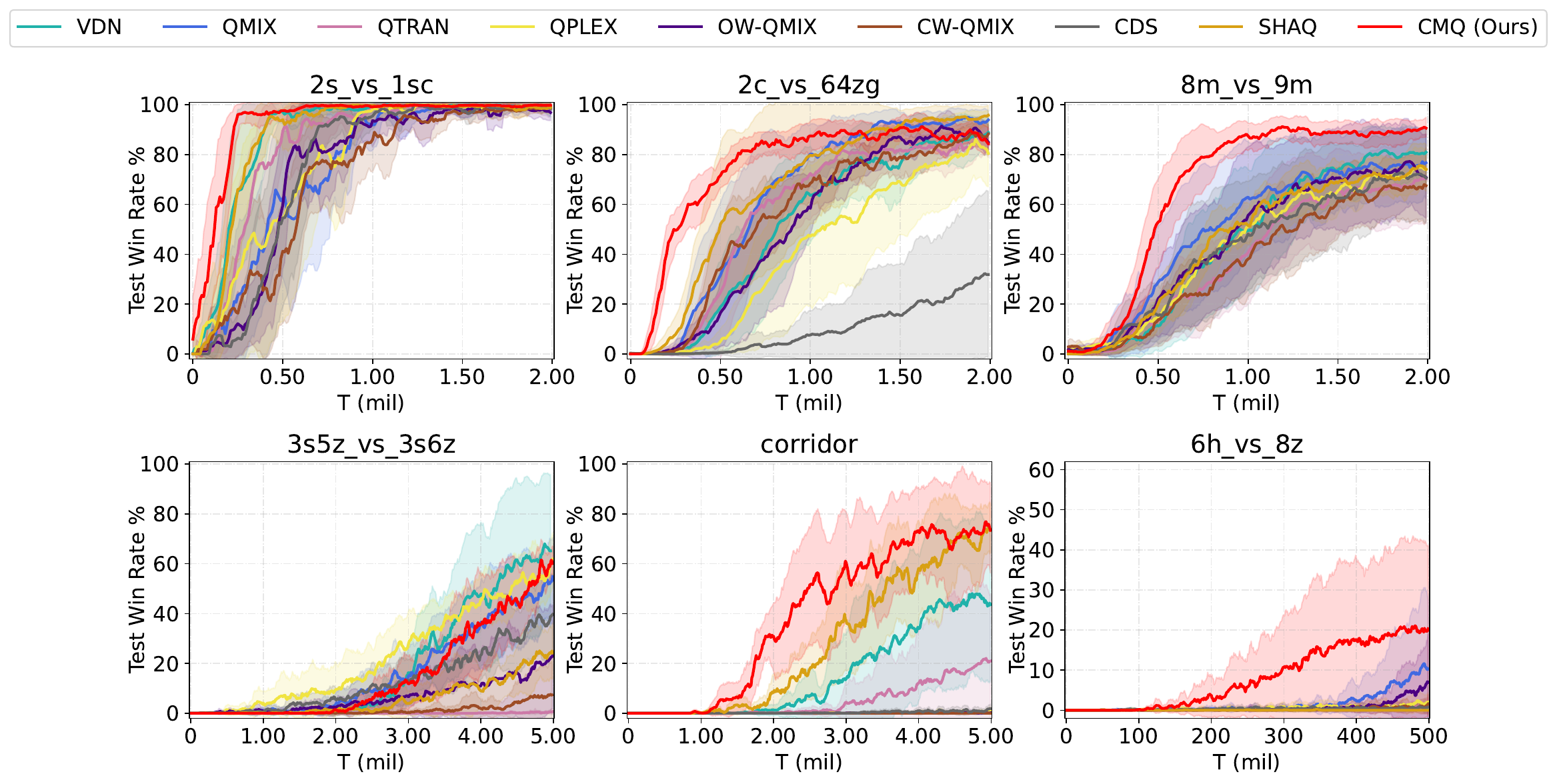}
    \caption{Performance comparison with baselines on easy and hard (first line), super hard (second line) scenarios.}
    \label{deepimage}
\end{figure*}

\begin{figure}[tb]
   \centering
    \includegraphics[width=1\columnwidth]{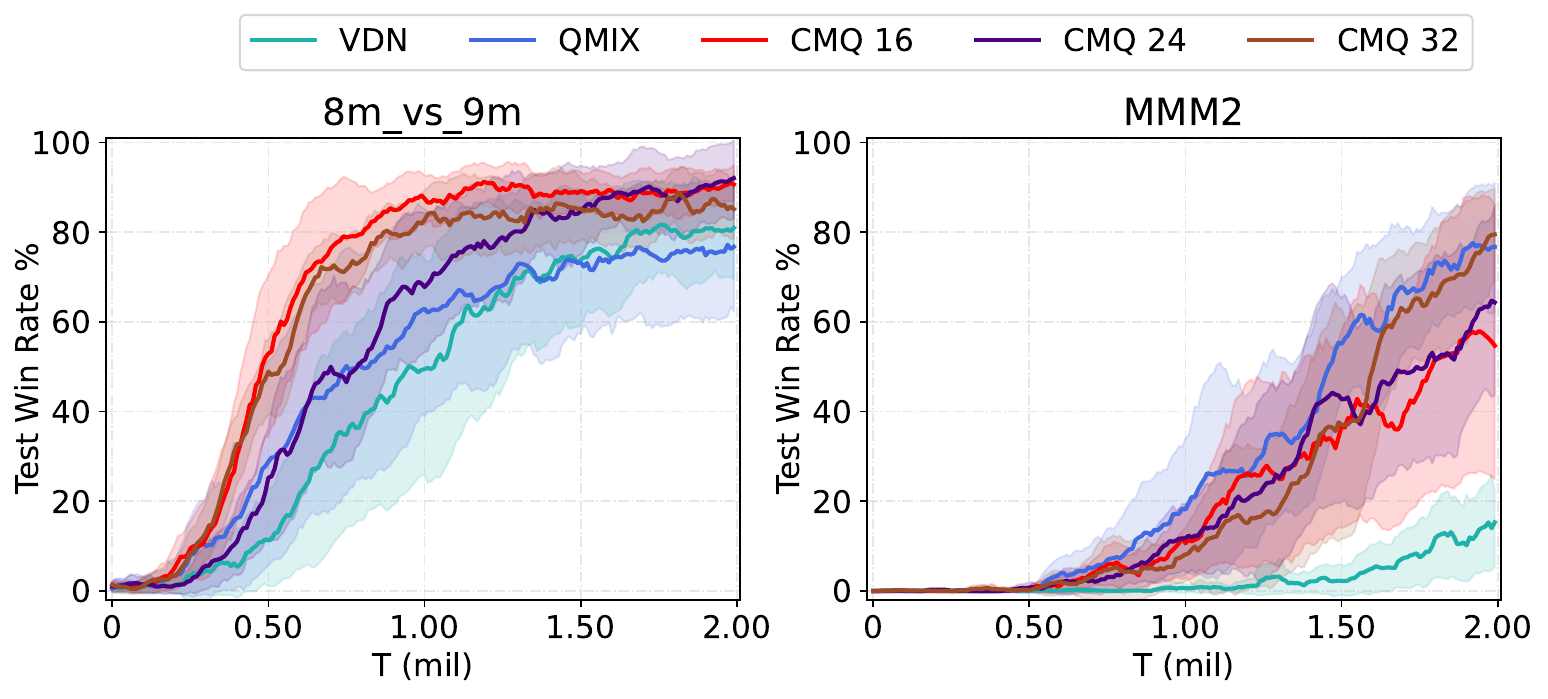}
    \caption{Performance with different number of concepts on 8m\_vs\_9m and MMM2 scenarios.}
    \label{parameter}
    \vskip -0.25in
\end{figure}

\subsection{Level Based Foraging}
We conducted experiments on two self-created LBF tasks within a $10\times10$ grid world where agents gather food, requiring cooperation at times.
Agents have a limited perception, seeing only a $5\times5$ area around them.
Cooperative actions are eating lower-level food, resulting in positive rewards, while non-cooperative moves result in a small penalty of $-0.002$.
The agents' actions include moving in four directions, eating, or doing nothing.
We assessed different algorithms by altering the number of agents and food amounts.

Figure~\ref{overview_results_lbf} shows the average test return curves of CMQ and baselines.
As the agent decreases, each agent needs more cooperative skills to find the food, and it takes more timesteps to obtain a high return.
CDS obtains a lower win rate, which may lack of diverse agents to explore collaborative strategies.
We find that CMQ obtains higher returns in $3$ agents with the $3$ food task.
It indicates that CMQ can better solve complex tasks with a more cooperative policy capacity, benefiting from the rich representation of cooperation concepts.

\subsection{SMAC Benchmark.}
Figure~\ref{deepimage} presents the performance of CMQ and baseline methods across various SMAC scenarios.
On easy maps like 2s\_vs\_1sc, CMQ performs on par with other algorithms, all of which are able to learn optimal policies. As the task difficulty increases, especially in maps requiring fine-grained coordination (e.g., 2s\_vs\_64zg and 8m\_vs\_9m), CMQ consistently outperforms existing methods. Notably, CMQ surpasses SHAQ and VDN by a small but consistent margin and significantly outperforms QMIX, QTRAN, QPLEX, and CDS. While VDN performs reasonably well in tasks with limited coordination demand, it fails to scale under complex cooperative settings. SHAQ benefits from its enhanced credit assignment but still lags behind CMQ, which leverages cooperation concepts to achieve more accurate decomposition and faster convergence.
On 8m\_vs\_9m, CMQ learns to execute focused fire and tactical retreats, outperforming all baselines by a large margin.

In super-hard scenarios, CMQ improves the average win rate of state-of-the-art methods by nearly 20\%. 
For instance, in the heterogeneous 6h\_vs\_8z map, agents must alternate kiting and attacking to succeed—a behavior that CMQ successfully learns, while others like VDN, QPLEX, and CDS nearly fail.
On corridor, CMQ achieves near-perfect results, while other algorithms struggle to even explore viable strategies. Although performance on 3s5z\_vs\_3s6z is not superior to all baselines, this may be due to the reduced need for complex credit assignment.
Still, CMQ shows faster convergence (within 1M steps) and maintains over 90\% win rate on maps that demand consistent coordination.
These results suggest that embedding cooperation concepts enhances both interpretability and policy quality, enabling CMQ to balance expressiveness and performance without added model complexity.

Further, we study the learning performance as the number of concepts increases.
As shown in Figure~\ref{parameter}, as the number of concepts increases, the learning performance is improved slightly, i.e., on the 8m\_vs\_9m map.
On MMM2 maps, the extra cooperation concept representation can improve the performance, indicating that a larger number of concepts is required.
It is consistent with our motivation that linear decomposition would not harm the expression of the model when given a providential concept representation space.
On the other hand, a larger number of concepts will lead to large computational costs.
Generally, a moderate number of concepts is enough for an appropriate trade-off between performance improvement and computation costs.

\begin{figure*}[ht]
   \centering
    \includegraphics[width=2\columnwidth]{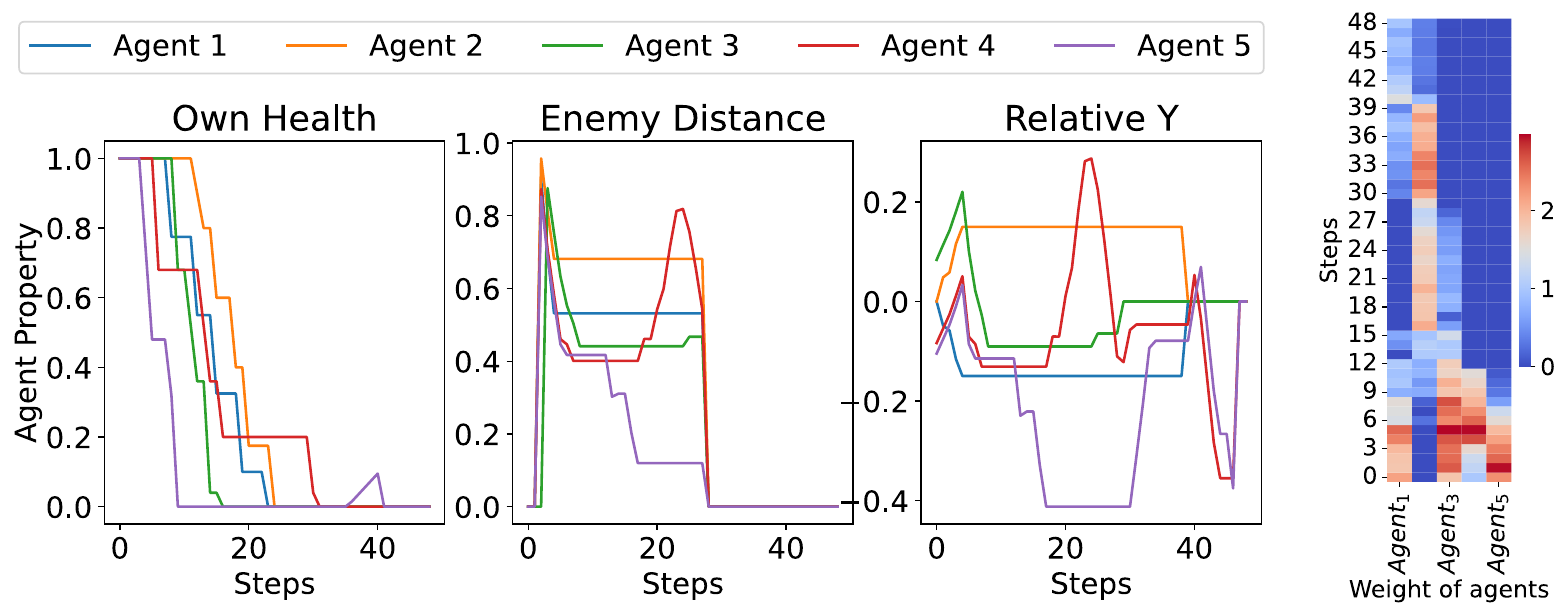}
    \caption{Visualisation of property semantics and agent contributions on the 2s3z scenario.}
    \label{interpretable1}
    \vskip -0.1in
\end{figure*}

\begin{figure*}[ht]
   \centering
    \includegraphics[width=1.6\columnwidth]{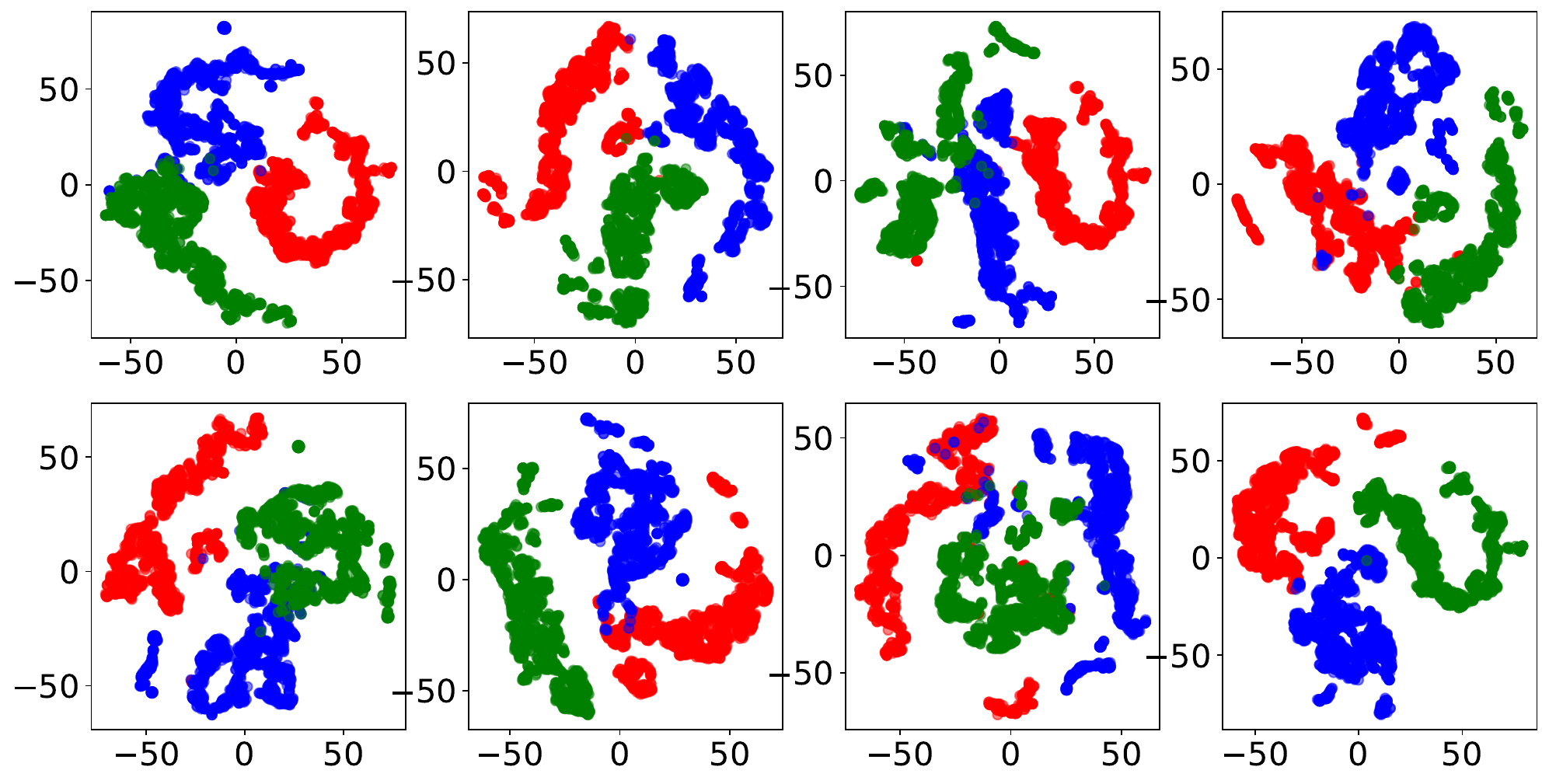}
    \caption{The t-SNE visualization of cooperative concepts learned of CMQ sample points for $30$ episodes, where three of all concepts are randomly selected to be combined into a group.}
    \label{interpretable2}
    \vskip -0.25in
\end{figure*}

\subsection{Interpretability}
To further demonstrate the interpretability of CMQ, we conduct an analysis on the 2s3z scenario using three agent-related features: own health, enemy distance, and relative Y-position.
We visualize one randomly selected active concept and the corresponding agent contributions over an episode (Figure~\ref{interpretable1}).
Notably, agent 5 shows high contribution early in the episode, as reflected by its strong activation in the heatmap—likely due to being attacked and experiencing a sharp drop in health.
By timestep $10$, agent 5's health reaches zero, and its contribution becomes strongly negative, indicating it has been eliminated.
In contrast, agent 4 maintains a low contribution in early stages due to its distance from enemies, reflected by minimal activation.
These findings illustrate how CMQ captures meaningful patterns in agent behavior, enabling attribution of action distributions to semantically grounded feature importance.

Beyond attribution, CMQ's cooperation concepts serve as rich and structured policy representations. 
To explore this qualitatively, we apply 2D t-SNE visualization to the embeddings of three randomly selected concepts on the 2s3z map (Figure~\ref{interpretable2}).
The results show clear clustering in the latent space: samples are grouped both by concept activation and by distinct cooperation modes within each concept.
This suggests that CMQ learns a latent hierarchy over cooperation semantics, enabling interpretable structure without supervision.
Such structured embeddings facilitate more informed exploration and contribute to CMQ's superior learning efficiency and generalization in complex tasks.

\section{Conclusion}
In this work, we propose CMQ to facilitate more efficient linear weighted value decomposition by incorporating concept learning into the mixer, which can maintain representational capability while enjoying interpretability.
Technically, we build the concept bottleneck via embedding the global state, which is used to deduce the contributions of each agent in value decomposition and predict the joint value function together with the individual value function.
The concept of bottleneck can incentivize our model to react positively to interventions at test time.
Empirical evaluations demonstrate that CMQ can achieve stable and efficient learning and help humans understand the cooperation behavior of MARL.
For future work, we aim to introduce concept learning into the entire pipeline of the MARL framework to yield further improvement and explore further theoretical properties of concept representation capacity in the MARL community.
Our results indicate that CMQ advances the state-of-the-art for the performance-vs-interpretability trade-off, which can make progress on a crucial concern in explainable AI.

\bibliographystyle{IEEEtran}
\bibliography{ijcai23}

\end{document}